\documentclass[letterpaper, 10 pt, conference]{ieeeconf}
\IEEEoverridecommandlockouts
\usepackage{graphicx}
\usepackage{float}
\usepackage{subfigure}

\usepackage{amssymb}
\usepackage{array}
\usepackage{amsmath}
\usepackage{url}
\usepackage{multirow}
\usepackage{booktabs}
\usepackage{threeparttable}
\usepackage{fixltx2e}
\usepackage{float}
\usepackage{booktabs}
\usepackage{graphicx}
\usepackage{diagbox}
\usepackage{mathtools}
\usepackage{amsmath}
\usepackage{bbding}
\usepackage{pifont}
\usepackage{wasysym}
\usepackage{tikz}
\usepackage{calc}
\usepackage{amssymb}
\usepackage{multirow}
\usepackage{rotating}
\usepackage{colortbl}
\usepackage{color}
\usepackage{ragged2e}
\usepackage[colorlinks,linkcolor=blue,anchorcolor=blue,citecolor=blue]{hyperref}
\usepackage{caption}
\usepackage{cite}
\usepackage{gensymb}
\usepackage[ruled,linesnumbered]{algorithm2e}

\begin{document}
%
\title{DANet: Density Adaptive Convolutional Network with Interactive Attention for 3D Point Clouds}
%
%
%

\author{Yong~He,
        Hongshan~Yu,
        Zhengeng~Yang,
        Wei Sun,
        Mingtao~Feng,
        Ajmal~Mian
\thanks{This work was partially supported by the National Natural Science Foundation of China (Grants 61973106, 62103137, U2013203 and U1913202), the Australian Research Council  Future Fellowship (No. FT210100268).}
\thanks{Hongshan Yu, Yong He, Zhengeng Yang, Wei Sun are with the National Engineering Laboratory for Robot Visual Perception and Control Technology, College of Electrical and Information Engineering, Hunan University, Lushan South Rd., Yuelu Dist., 410082, Changsha, China.}
\thanks{Mingtao Feng is with the School of Computer Science and Technology, Xidian University, Xi'an, 710071, China}
\thanks{Ajmal~Mian is with the Department of Computer Science, The University
of Western Australia, Perth, WA 6009, Australia}}

\markboth{IEEE ROBOTICS AND AUTOMATION LETTERS,~Vol.~x, 2022}%
{Shell \MakeLowercase{\textit{et al.}}: Bare Demo of IEEEtran.cls for IEEE Journals}

\maketitle

\begin{abstract}
Local features and contextual dependencies are crucial for 3D point cloud analysis. Many works have been devoted to designing better local convolutional kernels that exploit the contextual dependencies. However, current point convolutions lack robustness to varying point cloud density. Moreover, contextual modeling is dominated by non-local or self-attention models which are computationally expensive. To solve these problems, we propose density adaptive convolution, coined DAConv. The key idea is to adaptively learn the convolutional weights from geometric connections obtained from the point {\em density} and position. To extract precise context dependencies with fewer computations, we propose an interactive attention module (IAM) that embeds spatial information into channel attention along different spatial directions. DAConv and IAM are integrated in a hierarchical network architecture to achieve local density and contextual direction-aware learning for point cloud analysis. 
Experiments show that DAConv is significantly more robust to point density compared to existing methods and extensive comparisons on challenging 3D point cloud datasets show that our network achieves state-of-the-art classification results of 93.6\% on ModelNet40, competitive semantic segmentation results of 68.71\% mIoU on S3DIS and part segmentation results of 86.7\% mIoU on ShapeNet.
\end{abstract}

\begin{keywords}
Deep learning, 3D point clouds, Density adaptive convolution, Interactive attention module
\end{keywords}

\section{Introduction}
With the popularity of LiDAR scanners and depth cameras, 3D point clouds 
have become increasingly accessible, promoting a wide range of applications such as autonomous driving \cite{nunes2022segcontrast}, robotics \cite{chen2022direct}, and industrial automation \cite{li2022sim,hoang2022voting} etc. These sensors capture millions of points per second and hence, efficient processing becomes a critical issue. Early works transformed the irregular point clouds to regular grid representations such as multi-view images \cite{lawin2017deep}\cite{boulch2018snapnet} or voxels \cite{tchapmi2017segcloud},\cite{maturana2015voxnet}, so that the regular convolution kernels can be easily applied. However, such a pipeline sacrifices important geometric information and leads to unnecessary computational overhead during the representation transformation as well as subsequent learning and inference. \\ 
\indent A recent method that directly processes the point clouds spiked an increasing interest in the research community. The pioneering work PointNet \cite{qi2017pointnet} encodes the spatial information of point clouds by combining multi-layer perceptrons (MLPs) and global aggregation (e.g.maxpooling). Subsequent works\cite{qi2017pointnet++}\cite{liu2019point2sequence}\cite{li2018so} exploit local aggregation schemes to improve the network learning ability. Nevertheless, simple MLPs still treat each point individually and ignore the geometric connections between local/global points, and do not exploit the wider and finer local features as well as contextual dependencies. 

To exploit local features, many works get their inspiration from conventional 2D convolution and design similar convolutions for point clouds \cite{lei2020spherical}.
To process the unordered and irregular point clouds, some methods perform point convolutions \cite{li2018pointcnn}\cite{shen2018mining}\cite{wang2019dynamic}\cite{zhao2019pointweb} on the K-nearest neighbors of each point. However, such point convolutions are not robust to the varying density of point clouds \cite{hermosilla2018monte}. PointConv \cite{wu2019pointconv} uses a density function to re-weight the learned kernel weights, 
however, it still remains sensitive to the point cloud density. \\
\indent Another group of works\cite{thomas2019kpconv}\cite{hermosilla2018monte}\cite{Lei_2020_CVPR} performs point convolution on the ball neighborhood of a point that provides consistent metric information in the 3D space. These methods are still impacted by varying density.
MCC \cite{hermosilla2018monte} computes point density by Kernel Density Estimation (KDE) to adjust the kernels, where the convolution kernel is approximated by MLPs 
KPConv \cite{thomas2019kpconv} uses a linear correlation function of relative position between kernel points and neighbor points to predict the kernel weights, where the coefficients are determined offline manually according to the input point density. 
Overall, they also use the point density information to adjust the learned weights. However, besides requiring manual adjustment to different datasets, their re-weighting method overly complicates the  convolution operation, ignores the internal correlation between the point density and position. This method is still not sufficiently robust to varying density point clouds.

To exploit contextual dependency, early works aggregated multi-scale contextual features \cite{qi2017pointnet++} for high-level point feature learning. However, they still have limited receptive fields and fall short of fully exploring the global point dependencies. 
Recently, self-attention mechanism \cite{feng2020point}\cite{ma2020global} and non-local neural networks \cite{yan2020pointasnl} have been proposed for contextual feature learning given their high capacity to learn spatial or channel-wise dependencies. However, these methods require enormous computations. To increase the receptive field and depth of the network, some methods adopt a hierarchical architecture. However, the grouping in hierarchical networks can alter the global spatial information within groups, which is problematic for precise context learning. 

We propose Density Adaptive Convolution (DAConv) to learn robust local features from point clouds of varying density. The DAConv weights are treated as a continuous function. Although continuous convolution functions have been used before \cite{hermosilla2018monte}\cite{wu2019pointconv}, 
they adapt the same convolution design that associate coefficients with kernels
where the coefficients are learned from point density.
In contrast, the proposed DAConv has a weighting function that adaptively learns the convolutional weights from the points' geometric connections obtained from fusion of the point density and position. 
Directly learning the convolutional weighting function from the fusion of density and position not only simplifies the convolution operation but also enables the network to be highly robust to varying density. Since learning a large number of convolutional weights from limited geometric connections is inefficient, we reformulate the DAConv into an efficient version which divides convolutional weights into two parts, one learned by a weight function and the other learned in a data-driven manner. This improves the memory efficiency of DAConv without sacrificing robustness.

To efficiently explore precise contextual dependency, we propose an interactive attention module (IAM) that embeds precise spatial information into channel attention to learn direction-aware context. To elaborate, the IAM adopts two global pooling layers to respectively aggregate the input features along the group direction and the local direction into two direction-aware feature maps. The direction-aware feature maps are then encoded into two attention maps, 
to learn the long-range contextual dependencies along the group direction and the short-range contextual dependencies along the local direction. Finally, we integrate DAConv and IAM in a hierarchical network architecture for end to end training and subsequent inference.

To summarize, our contributions include: 
(1) DAConv, a new density adaptive convolution that 
is robust to point clouds of varying density. 
(2) An interactive attention module that efficiently learns short/long-range contextual dependencies to enhance point feature representation. 
(3) A hierarchical network architecture DANet that integrates DAConv and IAM for end to end point cloud processing. 
Our network is efficient and achieves state-of-the-art 93.6\% classification accuracy on ModelNet40 \cite{wu20153d}, and improves mIoU over the baselines by 13.4\% on S3DIS \cite{armeni20163d} and 3.9\% on ShapeNet \cite{yi2016scalable}. A control experiment on Modelnet40 shows that when the point cloud is downsampled from 1024 to only 64 points, the classification accuracy of DANet drops to 81.0\% whereas that of DGCNN, PointConv, PointNet and PointNet++ drop to 5.6\%, 18.4\%, 35.6\% and 73.4\%.

\vspace{-2mm}
\section{Related Work}\label{section2}
\vspace{-1mm}

\vspace{1mm}
\noindent \textbf{Point Convolution:} State-of-the-art deep neural networks directly process the raw point clouds to maximally preserve their geometric information. Pioneering work PointNet\cite{qi2017pointnet} uses shared MLPs to extract point-wise features and adopts a symmetric function such as max-pooling to collect these features into global features. Since max-pooling only captures the maximum activation across global points, PointNet cannot exploit local features that are crucial for vision tasks. 

\indent Follow up works proposed novel point convolutions on points or point graphs, similar to how convolution operation is performed on 2D images, albeit without the need for organizing the points on regular grids. One category of  methods\cite{xu2018spidercnn}\cite{wang2018deep}\cite{xu2021paconv}\cite{liu2019relation} group K-nearest neighbor points to perform convolution, where the convolutions approximate weight functions such as MLPs to learn the kernel weights. These methods lack robustness to varying density point clouds, since density directly influences the k-nearest neighbor search. 
PointConv \cite{wu2019pointconv} takes point density into account and uses the point position as input of the weight function, and employs another density function to re-weight the weighting function. However, this re-weighting neglects the internal correlation between point density and position. This deteriorates the network robustness and increases the computational complexity of convolution. 
These methods directly predict kernel weights to map input features to output features, which comes at a high computational cost and also increases the memory traffic during the learning process.

\indent Another line of works perform point convolution\cite{hermosilla2018monte}\cite{thomas2019kpconv} on radius neighborhood points to alleviate the impact of density. For example, MCC\cite{hermosilla2018monte} computes the point density by Kernel Density Estimation (KDE) to directly adjust convolution kernel weights, where the weight function is approximated by MLPs. This also has the same limitation as PointConv\cite{wu2019pointconv}. KPConv\cite{thomas2019kpconv} associates convolutional weights with a series of explicit kernel points in a local region, and predicts the kernel points weight through a linear correlation function of relative position between kernel points and neighbor points. The linear coefficient is chosen according to the input density to characterize the influence of density to kernel weights. However, the linear coefficient must be set offline i.e., the convolution needs to be specifically optimized for different datasets according to their point density. Unlike these density adaptive convolutions, the proposed DAConv directly learns kernel weights {\em online} from the {\em fusion of point density and position}. Moreover, it reduces the computational complexity and memory consumption of the weight function.

\begin{figure*}[htbp]
\centering
\includegraphics[width=7in]{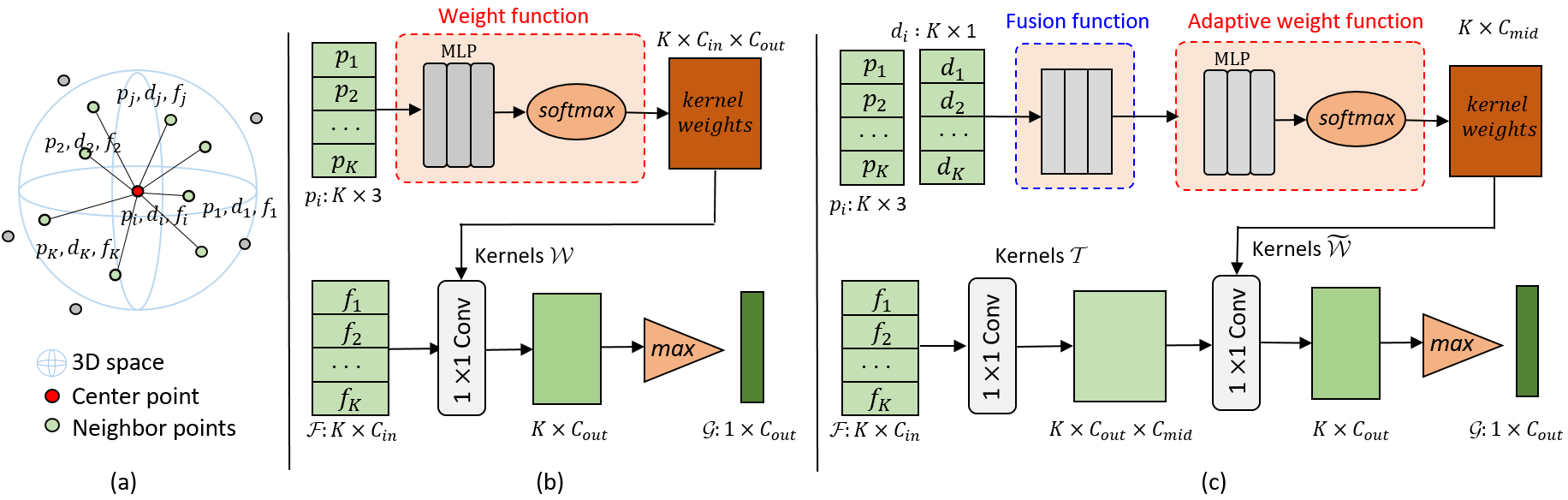}
\vspace{-2mm}
\caption{\textbf{DAConv:} Density Adaptive Convolution. (a) Local points with corresponding position $p$, density $d$ and features $f$; (b) General point convolution operation on one local region. (c) Proposed DAConv operating on a local region. DAConv reformulates the convolution kernels $\mathcal{W}$ into kernels $\mathcal{T}$ and kernels $\widetilde{\mathcal{W}}$, where the weight sizes are $C_{in}\times C_{mid}\times C_{out}$ and $K\times C_{mid}$ respectively.}
\vspace{-3mm}
\label{fig2}
\end{figure*}

\vspace{1mm}
\noindent \textbf{Contextual Dependency Learning:} Contextual dependency refers to spatial dependency e.g. to resolve the segmentation ambiguity of adjacent objects. There are four major contextual dependency learning strategies: multi-scale grouping\cite{engelmann2017exploring,qi2017pointnet++}, dilated mechanism\cite{engelmann2020dilated,komarichev2019cnn}, recurrent neural networks\cite{ye20183d,huang2018recurrent}  and attention mechanism.
Attention mechanism has become very popular for 2D image analysis. Motivated by its success in contextual information learning, attention mechanism has also been used in 3D point cloud processing. Especially, non-local/self attention networks are very popular in this domain due to their ability to build spatial or channel-wise attention. An example is AGCN\cite{xie2020point} which employs self-attention over local points to learn short-range point dependencies. Similarly, PAN\cite{feng2020point} employs self-attention over global points to learn long-range point dependencies, where the query points and key points are all global. On the other hand, PointASNL\cite{yan2020pointasnl} uses sampled points and global points as query and key points respectively, and employs a non-local mechanism to learn long-range contextual information. DGCNN-GCR\cite{ma2020global} employs self-attention on feature channel of global points to learn the long-range contextual features, all of which exploit non-local/ self-attention mechanisms to capture different types of spatial information. A downside of non-local/self-attention mechanisms is their high computational cost. We avoid this cost by talking an alternate approach.  We embed group and local spatial information into channel-attention to simultaneously explore long-range dependencies and short-range dependencies which is more efficient.

\vspace{-1mm}
\section{Method}
\label{section3}
\vspace{-1mm}
We first introduce the proposed DAConv 
followed by the interactive attention module (IAM) 
and finally 
introduce the overall architecture of our network.

\vspace{-2mm}
\subsection{DAConv: Density Adaptive Convolution}\label{section3_subsection1}
The proposed DAConv is inspired from the general point convolution. Hence, we first revisit the general definition of point convolution. 
Denoting the local points as $\mathcal{P}=\{p_{i}\mid i=1,2,\dots,K\}\in\mathbb{R}^{{K\times3}}$ (where $p_i$ is a point position vector) and their corresponding features as $\mathcal{F}=\{f_{i}\mid i=1,2,\dots,K\}\in\mathbb{R}^{{K\times C_{in}}}$.
In some cases, $p_i$ can also contain additional attributes such as color and surface normal. $K$ and $C_{in}$ are the number of local points and feature channels respectively. The general point convolution of $\mathcal{F}$ by kernels $\mathcal{W}$ at a point $p_i\in\mathbb{R}^{{N\times3}}$ can be formulated as
\begin{equation}\label{eq1}
\mathcal{G}= {\rm Conv}(\mathcal{W},\mathcal{F})=\sum_{j=1}^{K}\mathcal{W}(p_{j},p_{i})f_{j}, 
\end{equation}
where $\mathcal{W}$ is a weight function that learns kernel weights according to the relative position $(p_{j},p_{i})$ between the center point $p_i$ and its neighboring points $p_j$. $\sum$ refers to the aggregation function, `$\rm Conv$' indicates the general point convolution operation, and $\mathcal{G}\in\mathbb{R}^{{1 \times C_{out}}}$ are its output feature maps with $C_{out}$ channels.

Due to non-uniform sampling, point density varies over the global area. Intuitively, the contribution of dense points is different from sparse ones. 
We estimate the density $d_{i}$ at each point using kernel density estimation (KDE) as 
\begin{equation}
d_{i}=\frac{1}{K\sigma}\sum_{j\in{K}}G(\frac{p_{j}-p_{i}}{\sigma}), \label{eq2}
\end{equation}
where, $\sigma$ is the bandwidth that determines the smoothing of the resulting sample density function, $G(\cdot)$ is the density estimation kernel (we use a Gaussian).

Point density is derived from its position, and can thus be treated as one of the geometric connections of point clouds. Based on this finding, instead of employing the density to adjust the learned convolutional weights, we propose a simple but efficient density adaptive convolution (DAConv) that directly learns convolutional weights from the geometric connections obtained from the point density and position
\begin{equation}\label{eq3}
\mathcal{G}={\rm DAConv}(\mathcal{W},\mathcal{F})=\sum_{j=1}^{K}\mathcal{W}(L(p_{i},p_{j},d_{i},d_{j}))f_{j},
\end{equation}
where $d_{i}$ and $d_{j}$ denote the density of the center point and neighboring points, $p_{i}$ and $p_{j}$ denote their respective positions, and $L$ denotes the fusion function of density and position. 
Like most point convolutions, we approximate the weight function $\mathcal{W}$ as MLPs. However, the generation of weight matrix $\{\mathcal{W}(L(\cdot))|j=K\}\in\mathbb{R}^{{K\times C_{in}\times C_{out}}}$ requires large memory as well as learning a large number of weights from limited geometric connections which 
is inefficient.
Therefore, we reformulate an efficient version of DAConv based on the following lemma.\\
\noindent\textbf{Lemma:} \emph{DAConv is equivalent to the following formulation: $\mathcal{G}={\rm Max}({\rm Conv}_{1\times1}(\widetilde{\mathcal{W}},{\rm Conv}_{1\times1}(\mathcal{T},\mathcal{F})))$, where ${\rm Conv}_{1\times1}$ is 1$\times$1 convolution and Max is maxpooling, $\mathcal{T}\in\mathbb{R}^{C_{in}\times C_{mid}\times C_{out}}$ are the kernels of the first 1$\times$1 convolution, $\widetilde{\mathcal{W}}\in\mathbb{R}^{K\times C_{mid}}$ are the kernels of the second 1$\times$1 convolution.}\\
\noindent \textbf{Proof:} To better understand the DAConv reformulation, set $\{\mathcal{W}(j,c_{in})|j=K,c_{in}=C_{in}\}\in\mathbb{R}^{C_{out}}$ as a vector from weight matrix $\{\mathcal{W}(L(\cdot))|j=K\}\in\mathbb{R}^{{K\times C_{in}\times C_{out}}}$, and set $\mathcal{F}\in \mathbb{R}^{K\times C_{in}}$ as $\{\mathcal{F}(j,c_{in})|j=K,c_{in}=C_{in}\}$, where $j$ and $c_{in}$ are the index of the neighbor points and input feature channels. According to Eq.\ref{eq3}, DAConv can be expressed as
\begin{equation}\label{eq4}
\mathcal{G}={\rm DAConv}(\mathcal{W},\mathcal{F})=\sum_{j=1}^{K}\sum_{c_{in}=1}^{C_{in}}\mathcal{W}(j,c_{in})\mathcal{F}(j,c_{in}).
\end{equation}
Since the weight function is approximated by MLPs implemented as 1$\times$1 convolutions, the weight matrix generated by weight function can be expressed as
\begin{equation}\label{eq5}
\mathcal{W}(j,c_{in})=\sum_{c_{mid}=1}^{C_{mid}}\widetilde{\mathcal{W}}(j,c_{mid})\mathcal{T}^\mathrm{T}(c_{in},c_{mid}),
\end{equation}
where $c_{mid}$ and $C_{mid}$ are the index and number of output channels of middle layer, $\{\mathcal{T}(c_{in},c_{mid})|c_{in}=C_{in},c_{mid}=C_{mid}\}\in\mathbb{R}^{C_{out}}$ is a vector from $\mathcal{T}\in\mathbb{R}^{{C_{in}\times (C_{mid}\times C_{out})}}$. Substituting Eq.\ref{eq5} into Eq.\ref{eq4}, we get
\begin{equation}\label{eq6}
\begin{aligned}
\mathcal{G}&={\rm DAConv}(\mathcal{W},\mathcal{F})\\
&=\sum_{j=1}^{K}\sum_{c_{in}=1}^{C_{in}}(\sum_{c_{mid}=1}^{C_{mid}}\widetilde{\mathcal{W}}(j,c_{mid})\mathcal{T}^\mathrm{T}(c_{in},c_{mid}))\mathcal{F}(j,c_{in})\\
&= \sum_{j=1}^{K}\sum_{c_{mid}=1}^{C_{mid}}\widetilde{\mathcal{W}}(j,c_{mid})\sum_{c_{in}=1}^{C_{in}}(\mathcal{T}^\mathrm{T}(c_{in},c_{mid})\mathcal{F}(j,c_{in}))\\
&={\rm Max}({\rm Conv}_{1\times1}(\widetilde{\mathcal{W}},{\rm Conv}_{1\times1}(\mathcal{T},\mathcal{F}))).
\end{aligned}
\end{equation}

According to the above reformulation, DAConv comprises three operations including two 1$\times$1 convolutions and one maxpooling. Fig.~\ref{fig2} shows the DAConv operation on K-nearest neighbor points. 
Using this formulation, we divide the  1$\times$1 convolution kernels into two parts: convolution kernel $\mathcal{T}\in\mathbb{R}^{C_{in}\times C_{mid}\times C_{out}}$ and convolution kernel $\widetilde{\mathcal{W}}\in\mathbb{R}^{K\times C_{mid}}$. The complexity of the reformulated DAConv is $K\times C_{mid}+ C_{in}\times C_{mid}\times C_{out}$ compared to the original DAConv complexity of $K\times C_{in}\times C_{out}$. Under the setting $K=30$, $C_{mid}=16$ and $C_{in}=C_{out}=64$, the reformulated DAConv requires about 75\% less computational resources. 

The weight of kernels $\mathcal{T}$ are learned in a data driven manner, and the weights of kernels $\widetilde{\mathcal{W}}$ are dynamically learned though the adaptive weight function according to the geometric connection of the point clouds. The geometric relationship information is obtained by a fusion function according to the point density and position. 

\vspace{1mm}
\noindent \textbf{Fusion Function:} The weight function highly depends on the geometric connection of the input point clouds. Therefore, we construct sufficient geometric connections by fusing the density and position information.  The optimal fusion scheme is defined as 
\begin{equation}\label{eq7}
L(p_{i},p_{j},d_{i},d_{j}) = \left[p_{i},p_{j}-p_{i}, d_{j}-d_{i},||p_{j}-p_{i}||\right],
\end{equation}
where, $p_{j}-p_{i}$ is the position difference, $||p_{j}-p_{i}||$ is the 3D Euclidean distance, $d_{j}-d_{i}$ is the density difference, and $\left[,,\right]$ is the concatenation operation. 
Fusion enables online learning of the weights from multiple types of geometric connections. 
This not only makes DAConv more robust to varying density point clouds, but also simplifies the convolution operation, by avoiding setting additional density functions, leading to improved efficiency. 


\vspace{1mm}
\noindent \textbf{Adaptive Weight Function:} The goal of adaptive weight function is to learn the weights of kernels $\widetilde{\mathcal{W}}$. The outputs of weight function are
\begin{equation}\label{eq8}
\widetilde{\mathcal{W}}(L(p_{i},p_{j},d_{i},d_{j}))=\mathcal{S}(\phi(L(p_{i},p_{j},d_{i},d_{j})), \end{equation}
where $\phi$ is a non-linear function implemented with Multi-layer Perceptrons (MLPs). $\mathcal{S}$ indicates softmax normalization to keep the kernel weights in the range (0,1).


\subsection{Interactive Attention Module (IAM)}
\label{section3_subsection2}
The proposed IAM encodes precise spatial information of the points from group and local direction into feature channels. It then learns the dependencies between channels through channel attention. Hence, IAM learns to encode long/short-range spatial (contextual) dependencies. IAM is divided into two stages: spatial information encoding and spatial attention generation. Algorithm~\ref{algorithm} defines our IAM and its details are given below.

\vspace{1mm}
\noindent \textbf{Spatial Information Encoding:}  After the sampling and grouping operation, the global spatial information $\mathcal{F}\in\mathbb{R}^{n \times C}$ is disturbed and reorganized into group spatial information and local spatial information $\mathcal{F}\in\mathbb{R}^{N \times K \times C}$, where $n,N,K,C$ denote the number of global points, groups, local points and feature channels, respectively. Encoding only the global spatial information into feature channels cannot correctly preserve point spatial information. Therefore, we simultaneously encode spatial information from both the group direction and the local direction.

First, we use two global average pooling kernels $\mathcal{N}\in\mathbb{R}^{N\times1}$ and $\mathcal{K}\in\mathbb{R}^{1\times{K}}$ to encode the spatial information along the group direction and local direction into each feature channel, respectively. For clarity, we set $\mathcal{F}\in\mathbb{R}^{N \times K \times C}$ as $\{\mathcal{F}(i,j,c)|i=N,j=K,c=C\}$, where $i$, $j$, $c$ are the indices of the group, local point, and feature channel respectively. The output of the two average poolings can be formulated as
\vspace{-1mm}
\begin{equation}
\mathcal{F}_{mid}^N=\frac{1}{K} \sum_{j=1}^{K}\mathcal{F}(i,j,c), 
\end{equation}
\vspace{-2mm}
\begin{equation}
\mathcal{F}_{mid}^K=\frac{1}{N} \sum_{i=1}^{N}\mathcal{F}(i,j,c), 
\vspace{-1mm}
\end{equation}
which generate a pair of feature maps $\mathcal{F}_{mid}^N\in\mathbb{R}^{N\times 1\times C}$ and $\mathcal{F}_{mid}^K\in\mathbb{R}^{1\times K\times C}$. To further encode the connection between groups and local points, we concatenate the feature maps and pass them to a shared MLP that is implemented by a 1×1 convolution, yielding
\begin{equation}
\mathcal{F}_{out}^{NK}=\delta_{NK}(\alpha_{NK}([\mathcal{F}_{mid}^{N},\mathcal{F}_{mid}^{K}]))\textcolor{blue}{,}
\end{equation}
where $[,]$ is the concatenation operation along the spatial direction, $\delta_{NK}$ is a non-linear activation function, $\alpha_{NK}$ is an MLP, and $\mathcal{F}_{out}^{NK}\in\mathbb{R}^{1\times(N+K)\times{C}/r}$ is a feature map that encodes precise spatial information from both the group direction and local direction. Here, $r$ is the reduction ratio for controlling the computational complexity of encoding, discussed further in Section \ref{section4_subsection4}. Compared to the global point spatial information encoding, the proposed spatial information interactive encoding method not only preserves more precise spatial information but also decreases the computational burden.

\vspace{1mm}
\noindent \textbf{Spatial Attention Generation:} To learn spatial dependencies, we calculate the spatial attention maps. We split $\mathcal{F}_{out}^{NK}$ along the spatial directions into two separate tensors $\mathcal{F}_{out}^{N}\in\mathbb{R}^{1\times{N}\times{C}/r}$ and $\mathcal{F}_{out}^{K}\in\mathbb{R}^{1\times{K}\times{C}/r}$. The spatial attention maps along the group direction and local direction can be generated by separately transforming $\mathcal{F}_{out}^{N}$ and $\mathcal{F}_{out}^{K}$ to tensors with the same channel number $C$ to the input features as
\begin{equation}
\mathcal{A}^{N} = \delta_N(\alpha_N(\mathcal{F}_{out}^{N})),
\end{equation}
\begin{equation}
\mathcal{A}^{K} = \delta_K(\alpha_K(\mathcal{F}_{out}^{K})),
\end{equation}
where $\alpha_N$ and $\alpha_K$ are two channel attention mechanisms, which can be implemented with any differentiable architecture, we use multi-layer perceptron. $\delta_N, \delta_K$ are softmax normalizations that normalize the spatial dependencies to a group attention map $\mathcal{A}^{N}$ and a local point attention map $\mathcal{A}^{K}$ in the rang (0,1). The attention maps help IAM capture long-range dependencies and short-range dependencies, respectively. 

Finally, the input features $\mathcal{F}_{in}$ are transformed to new features by multiplying the two spatial attention weight vectors. The output features are added to the input features to obtain the final output $\mathcal{F}_{out}$. 
\begin{equation}
\mathcal{F}_{out} = \mathcal{A}^{N}*\mathcal{A}^{K}*\mathcal{F}_{in} + \mathcal{F}_{in}\textcolor{blue}{,}
\end{equation}
where $*$ represents element-wise product. IAM provides an opportunity to capture the long/short range spatial dependencies, which helps the network pay attention to the object of interest in the 3D space.

\begin{algorithm}
\caption{Interactive Attention Module }\label{algorithm}
\textcolor{blue}{\# \small $\mathcal{F}$ : input feature of IAM}\\
\textcolor{blue}{\# \small B: batch size, C: feature channel, N: group number}\\
\textcolor{blue}{\# \small K: local point number, r: redution ratio}\\
original\_features = $\mathcal{F}$ \textcolor{blue}{\small \# [B,C,N,K]}\\ 
B,C,N,K = $\mathcal{F}$.size()\\
$\mathcal{F}_{mid}^N$ = Avgpool2d($\mathcal{F}$, dim=3)\textcolor{blue}{ \small \#[B,C,N,1]}\\
$\mathcal{F}_{mid}^K$ = Avgpool2d($\mathcal{F}$, dim=2).permute(0,1,3,2)\textcolor{blue}{ \small \#[B,C,K,1]}\\

$\mathcal{F}_{out}^{NK}$ = concat([$\mathcal{F}_{mid}^N$,$\mathcal{F}_{mid}^K$] dim=2)\textcolor{blue}{ \small \#[B,C,N+K,1]}\\
features = MLP($\mathcal{F}_{out}^{NK}$)))\textcolor{blue}{ \small \# [B,C/r,N+K,1]}\\

$\mathcal{F}_{out}^{N}$,$\mathcal{F}_{out}^{K}$ = split(features, [N,K], dim=2)\textcolor{blue}{\# \small [B,C/r,N,1], [B,C/r,K,1]}\\
$\mathcal{F}_{out}^{K}$ = $\mathcal{F}_{out}^{K}$.permute(0,1,3,2)\textcolor{blue}{\small \# [B,C/r,1,K] }\\

$\mathcal{A}^{N}$ = softmax(MLP($\mathcal{F}_{out}^{N}$))\textcolor{blue}{\small \# [B,C,N,1]}\\
$\mathcal{A}^{K}$ = softmax(MLP($\mathcal{F}_{out}^{K}$))\textcolor{blue}{\small\#[B,C,1,K]}\\

output = product(original\_features, $\mathcal{A}^{N}$)\textcolor{blue}{\small\#[B,C,N,K]}\\
output = product(output, $\mathcal{A}^{K}$)\textcolor{blue}{\small\#[B,C,N,K]}\\
$\mathcal{F}_{out}$ = add(output, original\_features)\textcolor{blue}{\small\#[B,C,N,K]}\\
\end{algorithm}

\vspace{-2mm}
\subsection{Network Architecture}\label{section3_subsection4}
\vspace{-1mm}

Using the proposed DAConv and IAM modules, we design network architectures for classification and segmentation. We refer to it as DANet, i.e. Density Adaptive Network.  Encoding layers contains one sampling and grouping operation, optionally\footnote{For small datasets like ModelNet40, IAM is not required.} an IAM block and various DAConv blocks. Sampling and grouping are implemented with farthest point sampling (FPS) and K-nearest neighbor (KNN) search, respectively. Decoding layers contains one interpolating operation, various MLP and one DAConv. All encoding and decoding layers have batch normalization and leaky-ReLU activation. Here, we introduce several notations to explain our architectures in detail. $E_i(N_s,K,[C_1,..., C_d])$ presents the $i$-th encoding layer with $N_s$ center points and $K$ neighbor point search using  $d$ number of DAConv with feature dimension $C_j, j=1, 2, ..., d$. $D_i(K,[C_1, ..., C_{d-1},C_d])$ is the $i$-th decoding layer using $d-1$ number of MLPs and one DAConv with feature dimension $C_j,j=1,...,d$. The fully connected layer processes the features obtained from last encoder or decoder layer to predict the final scores. $FC(C_1,...,C_d)$ presents that the fully connected layer has $d$ layers with dimension $C_j, j= 1,2,...,d$, where the last dimension $C_d$ equals to the number of classes.  all networks are implemented in PyTorch and trained using two RTX 3090 GPUs.


\noindent\textbf{Classification Network:} The classification network includes three encoding layers and one fully connected layer. By above notations, these layers can be represented as: $EC_1 (256, ~32, [64, ~64, ~64])$, $EC_2  (64, ~32, [64, ~64, ~128])$, $EC_3  (none, all, [256, ~512, ~1024])$, $FC(1024, ~512, ~256, ~40)$. The bandwidths in DAConv in encoding layers are set to $(0.1, 0.2, none)$. We use a dropout ratio of 0.4 for training and the SGD optimizer with 0.9 momentum and 0.1 initial learning rate, which is reduced until 0.001 using cosine annealing.

\noindent\textbf{Segmentation Networks:} The semantic segmentation network has four encoding layers, four decoding layers and one fully connected layer. These layers can be expressed as:  $E_1(1024,~32,[32,~32,~64])$, $E_2 (256,~32,[64,~64,~128])$, $E_3 (64, ~32, [128,~128,~256])$, $E_4 (16,~32, [256, ~256,~512])$, $D_1 (256, 256)$,$D_2 (256, 256)$,$D_3 (256, 128)$,$D_4 (128, 128, 128)$, $FC(128, 128, 13)$. The DAConv bandwidths in encoding layers are set to $(0.1, 0.2, 0.4, 0.8)$.  
The {\bf part segmentation} network has three encoding and three decoding layers followed by a fully connected layer. These are represented as: 
$E_1 (512, ~32, [64, ~64, ~128])$, $E_2 (128, ~32, [128, ~128, ~256])$, $E_3 (none, all, [256, 512, 1024])$, $D_1(256, 256)$, $D_2 (256, 128)$, $D_3(128,128,128)$, $FC(128,128,50)$. The bandwidths in DAConv are set to $(0.1, 0.2, none)$. 
The semantic and part segmentation networks are both trained using 0.5 drop out ratio and the SGD optimizer with 0.9 momentum. The initial learning rate for semantic segmentation is set to 0.05 and that of part segmentation to 0.001. In both cases, the learning rate is reduced until 0.0001 using stepping. 

\section{Experimental Results}\label{section4}
We integrate DAConv and IAM into the PointNet++ architecture and evaluate our models for point cloud classification, semantic segmentation and part segmentation. Detailed network architectures and comparisons are provided as follows.

\begin{table}[t]
\centering
\caption{Classification results on ModelNet40 dataset. 
Para., PN, xyz, n, k and M stand for Parameters, PointNet, point coordinates, normal vector, thousand and million.}
\label{table1}
\scriptsize
\renewcommand\arraystretch{1.1}
\setlength{\tabcolsep}{1.8mm}{
\begin{tabular}{l|c|c|c|c|c}
\specialrule{1pt}{0pt}{0pt}
\textbf{Method}     & \textbf{Backbone} & \textbf{Input}    & \textbf{\#Points} &\textbf{Para.} & \textbf{OA}(\%) \\\hline
O-CNN\cite{wang2017cnn}      & Tree     & xyz, n & -     & -    & 90.6 \\
Kd-Net\cite{klokov2017escape}     & Tree     & xyz      & 32k   & -    & 91.8 \\
SO-Net\cite{li2018so}     & -        & xyz, n & 5k   & -     & 93.4 \\
PointNet\cite{qi2017pointnet}   & PN       & xyz      & 1k    & -    & 89.2 \\
KCNet\cite{shen2018mining}      & PN       & xyz      & 1k   & -     & 91.0 \\
SpiderCNN\cite{xu2018spidercnn}  & PN       & xyz, n & 5k   & -     & 92.4 \\
KPConv\cite{thomas2019kpconv}     & PN       & xyz      & 7k    & 6.15M    & 92.9 \\
DGCNN\cite{wang2019dynamic}      & PN       & xyz      & 1k    & -    & 92.9 \\ \hline
PointNet++\cite{qi2017pointnet++} & PN++    & xyz, n & 5k   & 1.48M     & 91.9 \\
SpecGCN\cite{wang2018local}    & PN++    & xyz      & 1k   & -     & 92.1 \\
PointWeb\cite{zhao2019pointweb}   & PN++    & xyz, n & 1k   & -     & 92.3 \\
PointConv\cite{wu2019pointconv}  & PN++    & xyz, n & 1k   & 1.96M     & 92.5 \\
A-CNN\cite{komarichev2019cnn}      & PN++    & xyz, n & 1k    & -    & 92.6 \\
PointASNL\cite{yan2020pointasnl}  & PN++    & xyz, n & 1k    & -    & 93.2 \\
RS-CNN\cite{liu2019relation}     & PN++       & xyz      & 1k   & 1.41M     & \textbf{93.6}  \\\hline
PointNet++\cite{qi2017pointnet++} & PN++    & xyz      & 1k   & -     & 90.7 \\
\textbf{DANet}       & PN++    & xyz      & 1k    & 1.37M    & \textbf{93.6}(2.9$\uparrow$) \\  \specialrule{1pt}{0pt}{0pt}
\end {tabular}}
\vspace{-3mm}
\end{table}


\vspace{-2mm}
\subsection{Classification}\label{section4_subsection1}
\vspace{-1mm}
\noindent \textbf{Data:} We evaluate our classification network on ModelNet40\cite{wu20153d} which comprises 12,311 CAD models from 40 categories. We use 9,843 models for training and 2,468 for testing. We uniformly sample 1024 points from each object and only use their xyz coordinates as input. During training, we augment the input data with random scaling in the range [0.67, 1.5], translation in the range [-0.2, 0.2], and shuffle the points. Similar to RS-CNN \cite{liu2019relation}, we perform 10 voting tests with random scaling and average the predictions during testing.


\vspace{1mm}
\noindent \textbf{Results:} Table \ref{table1} shows our results using the overall accuracy (OA) evaluation metric. For better comparison, we also show the backbone, input data type,  number of input points and model parameters for each network. As can be seen, our classification network  achieves the best accuracy (93.6\%) using only xyz point coordinates as input, which is a significant improvement of 2.9\% over the PointNet++. RS-CNN\cite{liu2019relation} also achieves similar results but with more less parameters. 

\begin{figure}[t]
\centering
\includegraphics[width= 2.8in]{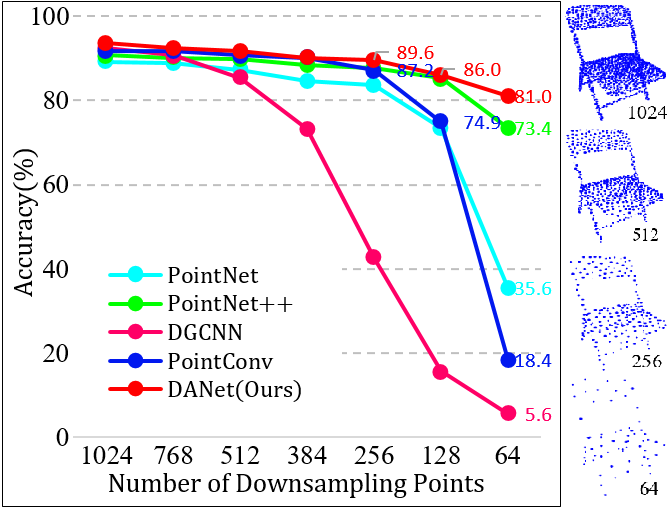}
\vspace{-1mm}
\caption{Comparison of classification results on ModelNet40 when points are downsampled. An example is shown on the right. DANet remains the most robust, even when there are only 64 points per object.}
\label{fig6}
\vspace{-4mm}
\end{figure}

\vspace{1mm}
\noindent \textbf{Robustness Analysis:} We compare the DANet's robustness to point density with several typical baselines 
PointNet \cite{qi2017pointnet}, PointNet++ \cite{qi2017pointnet++}, 
DGCNN \cite{wang2019dynamic}, and a classical point convolutional network  PointConv \cite{wu2019pointconv}.  For a fair comparison, all the networks are trained on modelnet40\_normal\_resampled dataset\cite{wu20153d} with 1024 points using only coordinates as the input. The test samples are downsampled to 1024, 768, 512, 384, 256, 128, 64 for inference by the model trained on 1024 points. Fig.\ref{fig6} shows that as input points get sparse, the classification accuracy of all networks drops, however, our DANet demonstrates significantly higher robustness compared to other networks. Interestingly, PointConv does not even perform as good as the PointNet++. 

\indent We further evaluate the robustness of our DANet to point permutation and grid transformation by augmenting the input data though rotation, translation, scaling and jittering at test time. Table \ref{table7} summarizes the results. As can been seen, all methods are invariant to permutation. DGCN, PointConv and DANet are sensitive to rotation. DGCN and PointConv are also sensitive to point scaling while PointNet++ and DANet are more robust. DANet achieves the best accuracy  under various transformations except 180\degree rotation.

\begin{table*}[h!]
\caption{Robustness to random point permutations and rigid transformations around $Y$-axis, translation along $Z$-axis, scaling 
and jittering. DANet achieves the best performance.}
\vspace{-3mm}
\label{table7}
\scriptsize
\centering
\renewcommand\arraystretch{1.1}
\setlength{\tabcolsep}{3mm}{
\begin{tabular}{l|c|c|c|c|c|c|c|c|c|c|c|c|c}
\specialrule{1pt}{3pt}{0pt}
\multirow{2}{*}{Methods} &\multicolumn{1}{c|}{\multirow{2}{*}{None}} &\multicolumn{1}{c|}{\multirow{2}{*}{Perm.}} & \multicolumn{3}{c|}{Rotation} & \multicolumn{2}{c|}{Translation} & \multicolumn{5}{c|}{Scaling}                     & \multirow{2}{*}{Jittering} \\ \cline{4-13}
 & \multicolumn{1}{c|}{}   &  \multicolumn{1}{c|}{}   & -90\degree  & 90\degree  & 180\degree  & +0.2   & -0.2   & 0.5-1.5 & 0.6-1.4 & 0.7-1.3 & 0.8-1.2 & 0.9-1.1 &  \\ \hline
PointNet++\cite{qi2017pointnet++}    & 92.1                  & 92.1                   & 57.9     & 57.9    & 57.9    & 90.7           & 90.8           & 91.2    & 91.2    & 90.9    & 91.0    & 91.0    & 8.96                    \\\hline
DGCNN\cite{wang2019dynamic}         & 92.5                  & 92.5                   & 55.6     & 56.5    & 74.0    & 92.3           & 92.3           & 90.7    & 91.6    & 92.1    & 92.3    & 91.8    & 91.5                    \\\hline
PointConv\cite{wu2019pointconv}                & 91.8                  & 91.8                   & 52.4     & 54.4    & \textbf{75.0}    & 91.8           & 91.8           & 84.4    & 87.5    & 89.9    & 90.6    & 91.0    & 90.6                    \\\hline
\textbf{DANet} (Ours)     & \textbf{93.0}     &  \textbf{93.0}       &  \textbf{58.7}     &  \textbf{59.3}    & 72.9    &  \textbf{92.6}   &  \textbf{92.7}  &  \textbf{92.3}    &  \textbf{92.4}    &  \textbf{92.3}    &  \textbf{92.6}    &  \textbf{92.6}    &  \textbf{91.6}  \\\specialrule{1pt}{0pt}{0pt}            
\end{tabular}
}
\vspace{-4mm}
\end{table*}

\vspace{-1mm}
\subsection{Semantic Segmentation}\label{section4_subsection2}
\noindent \textbf{Data:} We evaluate our semantic segmentation network on S3DIS\cite{armeni20163d} dataset which contains 3D RGB point clouds annotated with 13 classes, covering 271 rooms from 6 large-scale indoor scenes (total of 6020 square meters). During training, we uniformly sample 4096 points from each block with size 1m $\times$ 1m as input, and its corresponding feature as a 9-dimensional vector including coordinates, normalized color and normalized location in the room. Data augmentations consist of random rotation, scaling, and jittering. During testing, all points in each block are adopted. For extensive comparisons, we choose Area-5 as the test set which is not in the same building as the other areas.

\begin{table}[h]
\centering
\caption{Semantic segmentation results on S3DIS dataset Area-5. BLK and Grid indicate block and grid sampling, respectively, in data pre-processing. We report the mean class-wise intersection over union (mIoU) and mean class-wise accuracy (mAcc). }
\label{table2}
\scriptsize
\renewcommand\arraystretch{1.1}
\setlength{\tabcolsep}{1.9mm}{
\begin{tabular}{l|c|c|c}
\specialrule{1pt}{0pt}{0pt}
\textbf{Method}           & \textbf{Pre.} & \textbf{mAcc}  & \textbf{mIoU}  \\\hline
PointNet\cite{qi2017pointnet}   & BLK  & 48.98 & 41.09 \\
SPH3D-GCN\cite{lei2020spherical}  & Grid & 65.90  & 59.50  \\
SegCloud\cite{tchapmi2017segcloud}   & BLK  & 57.35 & 48.92   \\
PointCNN\cite{li2018pointcnn}   & BLK  & 63.86 & 57.26   \\
PCCN\cite{wang2018deep}       & BLK  & 67.01 & 58.27 \\
PointConv\cite{wu2019pointconv} & BLK  & 64.70 & 58.30   \\
PointWeb\cite{zhao2019pointweb}   & BLK  & 66.64 & 60.28 \\
KPConv\cite{thomas2019kpconv}     & Grid & 72.80  & 67.10    \\
PosPool\cite{liu2020closer}    & Grid & --    & 66.70    \\
SegGCN\cite{Lei_2020_CVPR}    & BLK  &70.40 &63.60 \\
PAConv\cite{xu2021paconv}     & BLK  & 73.00 & 66.58  \\\hline
PointNet++ (baseline)\cite{qi2017pointnet++}            & BLK  &62.85 & 53.37 \\
\textbf{DANet} &BLK   & \textbf{75.12} &\textbf{68.71}  \\\specialrule{1pt}{0pt}{0pt}
\end{tabular}
}
\vspace{-2mm}
\end{table}

\begin{figure}[!t]
\vspace{-4mm}
\centering
\includegraphics[width=3.5 in]{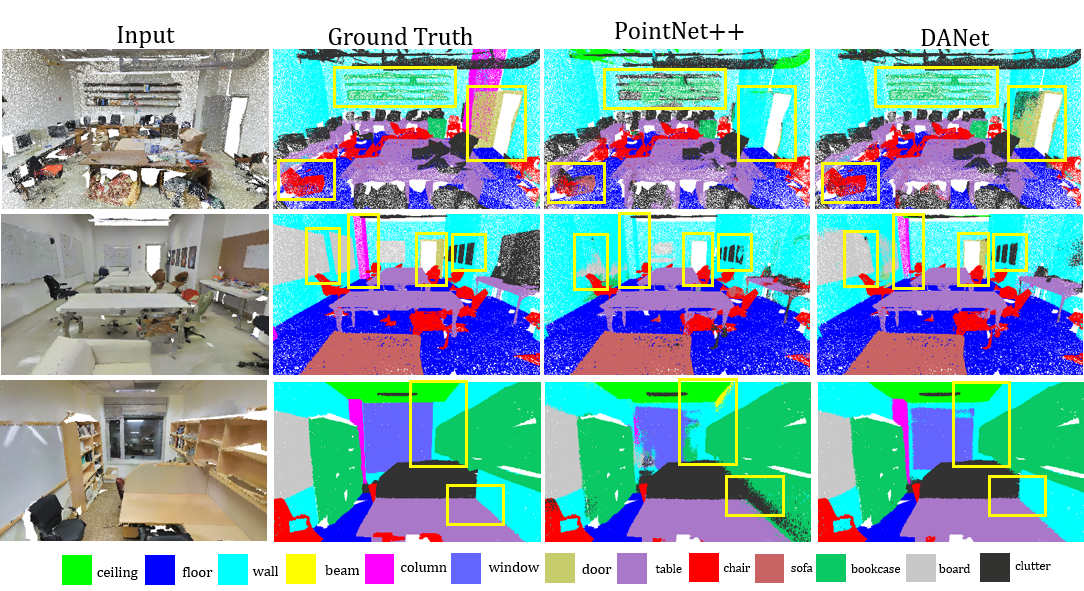}
\vspace{-6mm}
\caption{Visualization of semantic segmentation results on S3DIS Area-5. The yellow boxes highlight some examples in scenes where our method performs really well.}
\label{fig5}
\vspace{-2mm}
\end{figure}

\vspace{1mm}
\noindent \textbf{Results:} For evaluation, we use the mean class-wise intersection over union (mIoU) and the mean overall accuracy (mAcc) metrics. As shown in Table \ref{table2}, 
our DANet achieves the best mIoU of 68.71\% compared to various point convolution network, such as KPConv, PAConv.
DANet outperforms other approaches on 6 out of the 13 categories including ceiling, wall, column, table, sofa, and clutter. Furthermore, DANet achieves 10.41\% higher results compared to the classical density adaptive convolution PointConv.
We visualizations of semantic segmentation results are shown in Fig.\ref{fig5}.

\begin{table}[htbp]
\renewcommand\arraystretch{1.1}
\caption{Part segmentation results on ShapeNet Parts dataset. 
Our DANet achieves the highest class and instance mIoU and significantly improves over PointNet++.}
\label{table3}
\scriptsize
\centering
\setlength{\tabcolsep}{1.9mm}{
\begin{tabular}{l|c|c|c|c}
\specialrule{1pt}{0pt}{0pt}
\textbf{Method}   &\textbf{Input}  & \textbf{\#Points}   & \textbf{Cls.mIoU}(\%)  &\textbf{Ins.mIoU}(\%) \\\hline
Kd-Net\cite{klokov2017escape}       & xyz       & 4k      & 77.4    & 82.3 \\
SO-Net\cite{li2018so}               & xyz, n       & 1k      & 81.0    & 84.9 \\
KCNet\cite{shen2018mining}          & xyz       & 2k      & 82.2    & 84.7 \\
PointNet\cite{qi2017pointnet}       & xyz       & 2k      & 80.4    & 83.7 \\
PointNet++\cite{qi2017pointnet++}   & xyz       & 2k      & 81.9    & 85.1 \\
SpiderCNN\cite{xu2018spidercnn}     & xyz, n   & 2k      & 82.4    & 85.3 \\
SynspecCNN                          & mesh      & -       & 82.0    & 84.7 \\
PCNN\cite{wang2018deep}             & xyz       & 2k      & 81.8    & 85.1 \\
DGCNN\cite{wang2019dynamic}         & xyz       & 2k      & 82.2    & 85.1 \\
PointConv\cite{wu2019pointconv}     & xyz       & 2k      & 82.8    & 85.7           \\
KPConv\cite{thomas2019kpconv}       & xyz       & 2k      & 85.1    & 86.4 \\
RS-CNN\cite{liu2019relation}        & xyz       & 2k      & 84.0    & 86.2 \\\hline
PointNet++ (baseline)\cite{qi2017pointnet++}   & xyz, n       & 2k      & 81.9    & 85.1 \\
\textbf{DANet}            & xyz       & 2k      & 85.8 (\textbf{3.9}$\uparrow$) & 86.7 (\textbf{1.6}$\uparrow$)\\  \specialrule{1pt}{0pt}{0pt}
\end {tabular}}
\end{table}

\subsection{Part Segmentation}\label{section4_subsection3}
\noindent \textbf{Data:} We evaluate our network on ShapeNet Part segmentation \cite{yi2016scalable} dataset which contains 16881 shapes from 16 categories, 14006 for training and 2874 for testing. 2048 points are sampled from each object, and each object is point-wise labeled with 2 to 6 parts from a total of 50 parts.

\vspace{1mm}
\noindent \textbf{Results:} Table \ref{table3} shows our results using the Instance average mean Inter-over-Union (Ins.mIoU) and class average mean Inter-over-Union (Cls.mIoU) metrics. DANet improves the performance of baseline PointNet++ on Cls.mIoU by 2.1\% and Ins.mIoU by 0.3\%.

\subsection{Ablation Studies}\label{section4_subsection4}
We perform ablation studies on the S3DIS Area 5 \cite{armeni20163d} to evaluate the individual modules and performance under different reduction ratio $r$ in the IAM block. 

\noindent\textbf{Impact of DAConv and IAM:} Table \ref{tab:modules} shows the impact of our proposed modules. Replacing the MLPs in PointNet++ with DAConv improves the mIoU by 13.2\% points. 
Incorporating IAM into PointNet++ improves the mIoU by 6.54\%. 
This proves the effecacy of the individual modules. Integrating DAConv and IAM both into PointNet++ improves the mIoU by 15.34\% points. This shows the both modules can work in harmony to futher boost the performance.

\begin{table}[h]
\centering
\caption{Ablation Study: Effect of different modules on semantic segmentation accuracy (S3DIS Area-5 data).}
\label{tab:modules}
\scriptsize
\renewcommand\arraystretch{1.1}
\setlength{\tabcolsep}{4mm}{
\begin{tabular}{c|c|c|c|c}
\specialrule{1pt}{0pt}{0pt}
\textbf{MLPs} &\textbf{DAConv} &\textbf{IAM}   &\textbf{mAcc}    & \textbf{mIoU}(\%)   \\\hline
\checkmark & & &62.85 &53.37\\
&\checkmark& & 73.00     & 66.57           \\
\checkmark &&\checkmark & 67.24     &59.91           \\
&\checkmark&\checkmark  & {\bf 75.12}     &      {\bf 68.71}          \\
\specialrule{1pt}{0pt}{0pt}
\end{tabular}
}
\end{table}

\noindent\textbf{Impact of reduction ratio $r$:} We set different reduction ratios in IAM and report the changes in accuracy, FLOPS and model parameters in Table \ref{table6}. Increasing $r$ reduces FLOPs and network parameters marginally. More importantly, the accuracy (mIoU) remains relatively stable over different values of this parameter. 

\begin{table}[h]
\centering
\caption{Semantic segmentation results of DANet equipped with IAM blocks with  different reduction ratios $r$.}
\label{table6}
\scriptsize
\renewcommand\arraystretch{1.1}
\setlength{\tabcolsep}{3.5mm}{
\begin{tabular}{c|c|c|c}
\specialrule{1pt}{0pt}{0pt}
\textbf{Reduction ratio} $r$   &\textbf{mIoU}(\%)  & \textbf{FLOPs}  & \textbf{Parameters}  \\\hline
4  & 65.13  &  1224.7M    &  0.695M           \\
8   &  66.40   & 1221.5M &  0.662M              \\
16  &  \textbf{66.74}    & 1220.3M  &  0.647M\\
32   &  66.01 & \textbf{1219.9M}   &  \textbf{0.640M}          \\\specialrule{1pt}{0pt}{0pt}
\end{tabular}
}
\vspace{-2mm}
\end{table}

\section{Conclusion}
\label{sec_conclusion}
We proposed a novel Density Adaptive Convolutional Network (DANet) for 3D point cloud processing. Our network adaptively integrates local density-aware features and contextual direction-aware features. We proposed DAConv, a density adaptive convolution operation that operates on irregular 3D point clouds of varying densities to learn invariant features. DAConv adaptively learns the convolutional weights from the fusion of the point density and position. This enables the network to be highly robust to non-uniform sampling and at the same time, simplify the convolution operation. We reformulate the DAConv to an efficient version which greatly reduces the training complexity. We proposed IAM module that embeds precise spatial information into channel attention to simultaneously explore short-range contextual information along the local direction and long-range contextual information along the group direction. When DAConv and IAM are integrated into a hierarchical network architecture, DANet performs at par with state-of-the-arts methods, but with much higher robustness to point density and lower computational cost. Extensive experiments on challenging benchmarks, thorough ablation studies and theoretical analysis show the efficacy of the proposed modules.

\bibliography{reference}

\begin{thebibliography}{10}
\providecommand{\url}[1]{#1}
\csname url@samestyle\endcsname
\providecommand{\newblock}{\relax}
\providecommand{\bibinfo}[2]{#2}
\providecommand{\BIBentrySTDinterwordspacing}{\spaceskip=0pt\relax}
\providecommand{\BIBentryALTinterwordstretchfactor}{4}
\providecommand{\BIBentryALTinterwordspacing}{\spaceskip=\fontdimen2\font plus
\BIBentryALTinterwordstretchfactor\fontdimen3\font minus
  \fontdimen4\font\relax}
\providecommand{\BIBforeignlanguage}[2]{{%
\expandafter\ifx\csname l@#1\endcsname\relax
\typeout{** WARNING: IEEEtran.bst: No hyphenation pattern has been}%
\typeout{** loaded for the language `#1'. Using the pattern for}%
\typeout{** the default language instead.}%
\else
\language=\csname l@#1\endcsname
\fi
#2}}
\providecommand{\BIBdecl}{\relax}
\BIBdecl

\bibitem{nunes2022segcontrast}
L.~Nunes, R.~Marcuzzi, X.~Chen, J.~Behley, and C.~Stachniss, ``Segcontrast: 3d
  point cloud feature representation learning through self-supervised segment
  discrimination,'' \emph{IEEE Robotics and Automation Letters}, vol.~7, no.~2,
  pp. 2116--2123, 2022.

\bibitem{chen2022direct}
K.~Chen, B.~T. Lopez, A.-a. Agha-mohammadi, and A.~Mehta, ``Direct lidar
  odometry: Fast localization with dense point clouds,'' \emph{IEEE Robotics
  and Automation Letters}, vol.~7, no.~2, pp. 2000--2007, 2022.

\bibitem{li2022sim}
X.~Li, R.~Cao, Y.~Feng, K.~Chen, B.~Yang, C.-W. Fu, Y.~Li, Q.~Dou, Y.-H. Liu,
  and P.-A. Heng, ``A sim-to-real object recognition and localization framework
  for industrial robotic bin picking,'' \emph{IEEE Robotics and Automation
  Letters}, vol.~7, no.~2, pp. 3961--3968, 2022.

\bibitem{hoang2022voting}
D.-C. Hoang, J.~A. Stork, and T.~Stoyanov, ``Voting and attention-based pose
  relation learning for object pose estimation from 3d point clouds,''
  \emph{IEEE Robotics and Automation Letters}, vol.~7, no.~4, pp. 8980--8987,
  2022.

\bibitem{lawin2017deep}
F.~J. Lawin, M.~Danelljan, P.~Tosteberg, G.~Bhat, F.~S. Khan, and M.~Felsberg,
  ``Deep projective 3d semantic segmentation,'' in \emph{International
  Conference on Computer Analysis of Images and Patterns}.\hskip 1em plus 0.5em
  minus 0.4em\relax Springer, 2017, pp. 95--107.

\bibitem{boulch2018snapnet}
A.~Boulch, J.~Guerry, B.~Le~Saux, and N.~Audebert, ``Snapnet: 3d point cloud
  semantic labeling with 2d deep segmentation networks,'' \emph{Computers \&
  Graphics}, vol.~71, pp. 189--198, 2018.

\bibitem{tchapmi2017segcloud}
L.~Tchapmi, C.~Choy, I.~Armeni, J.~Gwak, and S.~Savarese, ``Segcloud: Semantic
  segmentation of 3d point clouds,'' in \emph{2017 international conference on
  3D vision (3DV)}.\hskip 1em plus 0.5em minus 0.4em\relax IEEE, 2017, pp.
  537--547.

\bibitem{maturana2015voxnet}
D.~Maturana and S.~Scherer, ``Voxnet: A 3d convolutional neural network for
  real-time object recognition,'' in \emph{2015 IEEE/RSJ International
  Conference on Intelligent Robots and Systems (IROS)}.\hskip 1em plus 0.5em
  minus 0.4em\relax IEEE, 2015, pp. 922--928.

\bibitem{qi2017pointnet}
C.~R. Qi, H.~Su, K.~Mo, and L.~J. Guibas, ``Pointnet: Deep learning on point
  sets for 3d classification and segmentation,'' in \emph{Proceedings of the
  IEEE conference on computer vision and pattern recognition}, 2017, pp.
  652--660.

\bibitem{qi2017pointnet++}
C.~R. Qi, L.~Yi, H.~Su, and L.~J. Guibas, ``Pointnet++ deep hierarchical
  feature learning on point sets in a metric space,'' in \emph{Proceedings of
  the 31st International Conference on Neural Information Processing Systems},
  2017, pp. 5105--5114.

\bibitem{liu2019point2sequence}
X.~Liu, Z.~Han, Y.-S. Liu, and M.~Zwicker, ``Point2sequence: Learning the shape
  representation of 3d point clouds with an attention-based sequence to
  sequence network,'' in \emph{Proceedings of the AAAI Conference on Artificial
  Intelligence}, vol.~33, no.~01, 2019, pp. 8778--8785.

\bibitem{li2018so}
J.~Li, B.~M. Chen, and G.~H. Lee, ``So-net: Self-organizing network for point
  cloud analysis,'' in \emph{Proceedings of the IEEE conference on computer
  vision and pattern recognition}, 2018, pp. 9397--9406.

\bibitem{lei2020spherical}
H.~Lei, N.~Akhtar, and A.~Mian, ``Spherical kernel for efficient graph
  convolution on 3d point clouds,'' \emph{IEEE transactions on pattern analysis
  and machine intelligence}, 2020.

\bibitem{li2018pointcnn}
Y.~Li, R.~Bu, M.~Sun, W.~Wu, X.~Di, and B.~Chen, ``Pointcnn: Convolution on
  x-transformed points,'' \emph{Advances in neural information processing
  systems}, vol.~31, pp. 820--830, 2018.

\bibitem{shen2018mining}
Y.~Shen, C.~Feng, Y.~Yang, and D.~Tian, ``Mining point cloud local structures
  by kernel correlation and graph pooling,'' in \emph{Proceedings of the IEEE
  conference on computer vision and pattern recognition}, 2018, pp. 4548--4557.

\bibitem{wang2019dynamic}
Y.~Wang, Y.~Sun, Z.~Liu, S.~E. Sarma, M.~M. Bronstein, and J.~M. Solomon,
  ``Dynamic graph cnn for learning on point clouds,'' \emph{Acm Transactions On
  Graphics (tog)}, vol.~38, no.~5, pp. 1--12, 2019.

\bibitem{zhao2019pointweb}
H.~Zhao, L.~Jiang, C.-W. Fu, and J.~Jia, ``Pointweb: Enhancing local
  neighborhood features for point cloud processing,'' in \emph{Proceedings of
  the IEEE/CVF Conference on Computer Vision and Pattern Recognition}, 2019,
  pp. 5565--5573.

\bibitem{hermosilla2018monte}
P.~Hermosilla, T.~Ritschel, P.-P. V{\'a}zquez, {\`A}.~Vinacua, and T.~Ropinski,
  ``Monte carlo convolution for learning on non-uniformly sampled point
  clouds,'' \emph{ACM Transactions on Graphics (TOG)}, vol.~37, no.~6, pp.
  1--12, 2018.

\bibitem{wu2019pointconv}
W.~Wu, Z.~Qi, and L.~Fuxin, ``Pointconv: Deep convolutional networks on 3d
  point clouds,'' in \emph{Proceedings of the IEEE/CVF Conference on Computer
  Vision and Pattern Recognition}, 2019, pp. 9621--9630.

\bibitem{thomas2019kpconv}
H.~Thomas, C.~R. Qi, J.-E. Deschaud, B.~Marcotegui, F.~Goulette, and L.~J.
  Guibas, ``Kpconv: Flexible and deformable convolution for point clouds,'' in
  \emph{Proceedings of the IEEE/CVF International Conference on Computer
  Vision}, 2019, pp. 6411--6420.

\bibitem{Lei_2020_CVPR}
H.~Lei, N.~Akhtar, and A.~Mian, ``Seggcn: Efficient 3d point cloud segmentation
  with fuzzy spherical kernel,'' in \emph{Proceedings of the IEEE/CVF
  Conference on Computer Vision and Pattern Recognition (CVPR)}, June 2020.

\bibitem{feng2020point}
M.~Feng, L.~Zhang, X.~Lin, S.~Z. Gilani, and A.~Mian, ``Point attention network
  for semantic segmentation of 3d point clouds,'' \emph{Pattern Recognition},
  vol. 107, p. 107446, 2020.

\bibitem{ma2020global}
Y.~Ma, Y.~Guo, H.~Liu, Y.~Lei, and G.~Wen, ``Global context reasoning for
  semantic segmentation of 3d point clouds,'' in \emph{Proceedings of the
  IEEE/CVF Winter Conference on Applications of Computer Vision}, 2020, pp.
  2931--2940.

\bibitem{yan2020pointasnl}
X.~Yan, C.~Zheng, Z.~Li, S.~Wang, and S.~Cui, ``Pointasnl: Robust point clouds
  processing using nonlocal neural networks with adaptive sampling,'' in
  \emph{Proceedings of the IEEE/CVF Conference on Computer Vision and Pattern
  Recognition}, 2020, pp. 5589--5598.

\bibitem{wu20153d}
Z.~Wu, S.~Song, A.~Khosla, F.~Yu, L.~Zhang, X.~Tang, and J.~Xiao, ``3d
  shapenets: A deep representation for volumetric shapes,'' in
  \emph{Proceedings of the IEEE conference on computer vision and pattern
  recognition}, 2015, pp. 1912--1920.

\bibitem{armeni20163d}
I.~Armeni, O.~Sener, A.~R. Zamir, H.~Jiang, I.~Brilakis, M.~Fischer, and
  S.~Savarese, ``3d semantic parsing of large-scale indoor spaces,'' in
  \emph{Proceedings of the IEEE Conference on Computer Vision and Pattern
  Recognition}, 2016, pp. 1534--1543.

\bibitem{yi2016scalable}
L.~Yi, V.~G. Kim, D.~Ceylan, I.-C. Shen, M.~Yan, H.~Su, C.~Lu, Q.~Huang,
  A.~Sheffer, and L.~Guibas, ``A scalable active framework for region
  annotation in 3d shape collections,'' \emph{ACM Transactions on Graphics
  (ToG)}, vol.~35, no.~6, pp. 1--12, 2016.

\bibitem{xu2018spidercnn}
Y.~Xu, T.~Fan, M.~Xu, L.~Zeng, and Y.~Qiao, ``Spidercnn: Deep learning on point
  sets with parameterized convolutional filters,'' in \emph{Proceedings of the
  European Conference on Computer Vision (ECCV)}, 2018, pp. 87--102.

\bibitem{wang2018deep}
S.~Wang, S.~Suo, W.-C. Ma, A.~Pokrovsky, and R.~Urtasun, ``Deep parametric
  continuous convolutional neural networks,'' in \emph{Proceedings of the IEEE
  Conference on Computer Vision and Pattern Recognition}, 2018, pp. 2589--2597.

\bibitem{xu2021paconv}
M.~Xu, R.~Ding, H.~Zhao, and X.~Qi, ``Paconv: Position adaptive convolution
  with dynamic kernel assembling on point clouds,'' in \emph{Proceedings of the
  IEEE/CVF Conference on Computer Vision and Pattern Recognition}, 2021, pp.
  3173--3182.

\bibitem{liu2019relation}
Y.~Liu, B.~Fan, S.~Xiang, and C.~Pan, ``Relation-shape convolutional neural
  network for point cloud analysis,'' in \emph{Proceedings of the IEEE/CVF
  Conference on Computer Vision and Pattern Recognition}, 2019, pp. 8895--8904.

\bibitem{engelmann2017exploring}
F.~Engelmann, T.~Kontogianni, A.~Hermans, and B.~Leibe, ``Exploring spatial
  context for 3d semantic segmentation of point clouds,'' in \emph{Proceedings
  of the IEEE international conference on computer vision workshops}, 2017, pp.
  716--724.

\bibitem{engelmann2020dilated}
F.~Engelmann, T.~Kontogianni, and B.~Leibe, ``Dilated point convolutions: On
  the receptive field size of point convolutions on 3d point clouds,'' in
  \emph{2020 IEEE International Conference on Robotics and Automation
  (ICRA)}.\hskip 1em plus 0.5em minus 0.4em\relax IEEE, 2020, pp. 9463--9469.

\bibitem{komarichev2019cnn}
A.~Komarichev, Z.~Zhong, and J.~Hua, ``A-cnn: Annularly convolutional neural
  networks on point clouds,'' in \emph{Proceedings of the IEEE/CVF Conference
  on Computer Vision and Pattern Recognition}, 2019, pp. 7421--7430.

\bibitem{ye20183d}
X.~Ye, J.~Li, H.~Huang, L.~Du, and X.~Zhang, ``3d recurrent neural networks
  with context fusion for point cloud semantic segmentation,'' in
  \emph{Proceedings of the European Conference on Computer Vision (ECCV)},
  2018, pp. 403--417.

\bibitem{huang2018recurrent}
Q.~Huang, W.~Wang, and U.~Neumann, ``Recurrent slice networks for 3d
  segmentation of point clouds,'' in \emph{Proceedings of the IEEE Conference
  on Computer Vision and Pattern Recognition}, 2018, pp. 2626--2635.

\bibitem{xie2020point}
Z.~Xie, J.~Chen, and B.~Peng, ``Point clouds learning with attention-based
  graph convolution networks,'' \emph{Neurocomputing}, vol. 402, pp. 245--255,
  2020.

\bibitem{wang2017cnn}
P.-S. Wang, Y.~Liu, Y.-X. Guo, C.-Y. Sun, and X.~Tong, ``O-cnn: Octree-based
  convolutional neural networks for 3d shape analysis,'' \emph{ACM Transactions
  On Graphics (TOG)}, vol.~36, no.~4, pp. 1--11, 2017.

\bibitem{klokov2017escape}
R.~Klokov and V.~Lempitsky, ``Escape from cells: Deep kd-networks for the
  recognition of 3d point cloud models,'' in \emph{Proceedings of the IEEE
  International Conference on Computer Vision}, 2017, pp. 863--872.

\bibitem{wang2018local}
C.~Wang, B.~Samari, and K.~Siddiqi, ``Local spectral graph convolution for
  point set feature learning,'' in \emph{Proceedings of the European conference
  on computer vision (ECCV)}, 2018, pp. 52--66.

\bibitem{liu2020closer}
Z.~Liu, H.~Hu, Y.~Cao, Z.~Zhang, and X.~Tong, ``A closer look at local
  aggregation operators in point cloud analysis,'' in \emph{European Conference
  on Computer Vision}.\hskip 1em plus 0.5em minus 0.4em\relax Springer, 2020,
  pp. 326--342.

\end{thebibliography}

\end{document}